\documentclass[a4paper]{scrartcl}
\usepackage[utf8]{inputenc}
\usepackage[T1]{fontenc}
\usepackage[english]{babel} 
\usepackage{lmodern}
\usepackage[]{algorithm2e}
\usepackage[onehalfspacing]{setspace}
\usepackage{amsmath,amssymb,amsthm,amsfonts,amsbsy,latexsym}
\usepackage{graphicx}
\usepackage{geometry}
\usepackage{natbib}
\usepackage{multirow}
\usepackage{float}
\usepackage{fancyhdr}
\usepackage{makecell}
\usepackage{pgf,tikz}
\usepackage{makeidx}
\usetikzlibrary{shapes,snakes}
\usetikzlibrary{fadings}
\usepackage[pdfpagelabels=true]{hyperref}

\newcommand{\E}{\mbox{I\negthinspace E}}

\newcommand{\R}{\mathbb{R}}

\newcommand{\N}{\mathbb{N}}

\newcommand{\D}{\mathcal{D}}

\DeclareMathOperator*{\argmin}{argmin}

\DeclareMathOperator*{\SNR}{SNR}

\DeclareMathOperator*{\Var}{Var}

\numberwithin{equation}{section}
\IfFileExists{upquote.sty}{\usepackage{upquote}}{}
\begin{document}

\newtheorem{thm}{Theorem}[section]
\newtheorem{Def}{Definition}[section]
\newtheorem{lem}{Lemma}[section]
\newtheorem{rem}{Remark}[section]
\newtheorem{cor}{Corollary}[section]
\newtheorem{ex}{Example}[section]
\newtheorem{ass}{Assumption}[section]
\newtheorem*{bew}{Proof}
\title{Loss-guided Stability Selection}
\author{Tino Werner\footnote{Institute for Mathematics, Carl von Ossietzky University Oldenburg, P/O Box 2503, 26111 Oldenburg (Oldb), Germany, \texttt{tino.werner1@uni-oldenburg.de}}}
\maketitle

\begin{footnotesize} 

\begin{abstract}

In modern data analysis, sparse model selection becomes inevitable once the number of predictors variables is very high. It is well-known that model selection procedures like the Lasso or Boosting tend to overfit on real data. The celebrated Stability Selection overcomes these weaknesses by aggregating models, based on subsamples of the training data, followed by choosing a stable predictor set which is usually much sparser than the predictor sets from the raw models. The standard Stability Selection is based on a global criterion, namely the per-family error rate, while additionally requiring expert knowledge to suitably configure the hyperparameters. Since model selection depends on the loss function, i.e., predictor sets selected w.r.t. some particular loss function differ from those selected w.r.t. some other loss function, we propose a Stability Selection variant which respects the chosen loss function via an additional validation step based on out-of-sample validation data, optionally enhanced with an exhaustive search strategy. Our Stability Selection variants are widely applicable and user-friendly. Moreover, our Stability Selection variants can avoid the issue of severe underfitting which affects the original Stability Selection for noisy high-dimensional data, so our priority is not to avoid false positives at all costs but to result in a sparse stable model with which one can make predictions. Experiments where we consider both regression and binary classification and where we use Boosting as model selection algorithm reveal a significant precision improvement compared to raw Boosting models while not suffering from any of the mentioned issues of the original Stability Selection.

\end{abstract}

\end{footnotesize}

\section{Introduction} 

The first milestone in high-dimensional data analysis has been achieved once the Lasso (\cite{tibsh96}) was introduced, performing automated sparse variable selection for the squared loss and a shrinking of the coefficients. Several years later, B\"{u}hlmann and Yu proposed $L_2$-Boosting (\cite{bu03}, see also \cite{bu06}), a sophisticated forward step-wise approach that combines simple linear regression models which turned out to be competitive to the Lasso, see for example \cite{efron04}, \cite{bu07}, \cite{bu06} for experiments and discussions on the differences of Lasso and $L_2$-Boosting. 

However, although both algorithms have attractive asymptotic properties (\cite{bu06}, \cite{bu}), they still show an overfitting behaviour in practice. This issue resulted in several modifications of the Lasso, for example, the Adaptive Lasso, which is a two-stage procedure (\cite{zou06}) where the second stage thins out the selected set of variables from the first stage, or the Multistep Adaptive Lasso, originally introduced in \cite{bu08}, which is an extension of the Adaptive Lasso and asymptotically performs $\log$-regularized least squares minimization. 

As for Boosting, the variant Sparse Boosting (\cite{bu06a}) has been proposed which respects the model complexity, measured by the trace of the corresponding Boosting operator, leading to sparser models. Nevertheless, a standard approach to prevent Boosting models from overfitting is a suitable choice of the number of iterations. Since iterating the Boosting algorithm as long as convergence occurs is not reasonable in terms of sparse variable selection, one tries to stop the algorithm before convergence. For this method, called ''early stopping'', there already has been done a lot of work, see for example \cite{bu03}, \cite{zhang05b}, \cite{bu06} and \cite{mayr12}. Note that \cite{hofner15} point out that even Boosting with early stopping still tends to overfitting. 

The analogous problem of finding the optimal number of iterations in Boosting models is to define a suitably chosen regularization parameter for Lasso-type models. Although cross-validation is a standard approach, \cite{bu10} point out that even for a sophisticatedly chosen grid $\Lambda$ for the regularization parameter, the optimal model may not be contained in the set $\{\hat S(\lambda) \ |\ \lambda \in \Lambda\}$ if $\hat S(\lambda)$ is the set of variables chosen by the Lasso with regularization parameter $\lambda$. They suggest a very fruitful strategy called ''Stability Selection'' which essentially combines the variable selection procedure with bagging (\cite{breiman96}) or, more precisely, subagging (\cite{bu02}), by aggregating models computed on subsamples. Only those variables that have been selected on sufficiently many subsamples enter the final stable model. The reason behind this model aggregation procedure is to immunize the final model against peculiar sample configurations. Stability Selection is a very powerful method that can be applied to Lasso or Graphical Lasso models (\cite{bu10}), but which also has been extended to Boosting models in \cite{hofner15}. The Stability Selection has been applied successfully in the context of gene expression (\cite{bu10}, \cite{stekhoven}, \cite{shah13}, \cite{hofner15}), fMRI data (\cite{ryali}) and voice activity detection (\cite{hamaidi}), which are exemplary for having very few observations with a huge number of predictors. 

Despite the magnificent success of Stability Selection, there are still open questions. The Lasso resp. Boosting models are computed on different (training) subsamples and these models are aggregated without ever performing an out-of-sample validation, which is known to be potentially misleading. Second, the choice of the tuning parameters of the Stability Selection may require expert knowledge to be defined appropriately. We experimentally show that, on noisy high-dimensional data, the error control in terms of the expected number of falsely selected variables to which both parameters are related (\cite[Thm. 1]{bu10}) seems to be too strict, may resulting in an empty model. Third, variable selection clearly depends on the loss function in the sense that model selection w.r.t. different loss functions ends up in different models which is not represented by the global per-family error rate criterion. 

Therefore, we propose a loss-guided Stability Selection variant that includes a validation step after having aggregated the subsample-specific models. In this validation step, we compare the performance w.r.t. the chosen loss function for different stable candidate models on an out-of-sample validation data set. The candidate stable models are computed using a pre-defined grid of either thresholds, i.e., the stable models include the variables whose aggregated selection frequencies exceed the respective threshold, or of cardinalities, i.e., the stable models consist of the respective number of variables with the highest aggregated selection frequencies. Then, the final stable model is the best performing candidate stable model. Thanks to the sparseness of the candidate stable models, this validation step induces only very little computational overhead compared to the original Stability Selection. Motivated by the issue that, on very noisy data, even the relevant variables may be selected on a rather low fraction of the models so that in may even happen that they achieve even lower selection frequencies than some noise variables, we propose another strategy where the grid search is replaced by an exhaustive search over the best variables, again with a limited computational overhead. Summarizing, we focus on finding a stable model with which one can make predictions instead of concentrating on avoiding false positives.

This article is organized as follows. In Sec. \ref{prelimsec}, we recapitulate Gradient Boosting and strategies that have been proposed to improve the model selection quality on real data and briefly summarize how Stability Selection works. Sec. \ref{ourstabselsec} is devoted to our Stability Selection variants. In Sec. \ref{simsec}, we conduct experiments, both on simulated and real data, that show the potential of our Stability Selection variant. Sec. \ref{outlook} concludes.

\section{Preliminaries} \label{prelimsec}

\subsection{Boosting and variable selection} \label{boostselsec} 

Let $\D=(X,Y)$ be a data set where $X \in \R^{n \times p}$ is the predictor matrix and where $Y \in \R^n$ is the corresponding response column. The rows of the regressor matrix $X_i$ are contained in some regressor space $\mathcal{X}\subset \R^p$, the responses $Y_i$ belong to some space $\mathcal{Y}\subset \R$. We assume that the instances $(X_i,Y_i)$ are i.i.d. realizations of some underlying joint distribution. We denote the submatrix of $X$ with all column indices in some index set $J$ as $X_{\cdot,J}$. Let $L$ be a loss function which in this work is a map $L: \mathcal{Y} \times \mathcal{Y} \rightarrow \R_{\ge 0}$. 
 
The idea behind Boosting is to combine simple models (''weak learners'', ''baselearners'') that are easy to fit in order to generate a final ''strong'' learning model. The first algorithm of this kind was the \texttt{Adaboost} algorithm (\cite{freund97}) for binary classification problems. It has been shown in \cite{breiman99} that \texttt{AdaBoost} can be seen as a gradient descent algorithm for the exponential loss. \cite{friedman} showed that \texttt{AdaBoost} can also be identified with a forward stage-wise procedure. We recapitulate the general \textbf{functional gradient descent (FGD)} algorithm in Alg. \ref{funcBoosting} (cf. \cite{bu07}).\ \\

\begin{algorithm}[H] 
\label{funcBoosting}
\textbf{Initialization:} Data $(X,Y)$, step size $\kappa \in ]0,1]$, number $m_{iter}$ of iterations and the offset value $\hat f^{(0)}(\cdot)\equiv \argmin_c\left(\frac{1}{n}\sum_i L(Y_i,c)\right)$\;
\For{$k=1,...,m_{iter}$}{
Compute the negative gradients $U_i=-\partial_f L(Y_i,f)|_{f=\hat f^{(k-1)}(X_i)}$ and evaluate them at the current model for all $i=1,...,n$\;
Treat the vector $U=(U_i)_i$ as response and fit a model $(X_i,U_i)_i \overset{\text{base procedure}}{\longrightarrow} \hat g^{(k)}(\cdot)$ with a pre-selected learning algorithm as base procedure \;
Update the current model via $\hat f^{(k)}(\cdot)=\hat f^{(k-1)}(\cdot)+\kappa \hat g^{(k)}(\cdot)$
}
\caption{Generic functional Gradient Boosting} 
\end{algorithm}\

The weak learners in the ``base procedure'' can, for example, be trees, smoothing splines or simple least squares models. The special case where the loss function is the squared loss is referred to as ``$L_2$-Boosting'' in this article. For an overview on Gradient Boosting algorithms and their paradigms, we refer to \cite{bu03} and \cite{bu07}.

The number $m_{iter}$ of Boosting iterations becomes very important if a pure Boosting algorithm is applied without further variable selection criteria due to a general overfitting behaviour of Boosting (see e.g. \cite{bu07}). More precisely, a small number of $m_{iter}$ leads to biased models with a low variance whereas performing many iterations reduces the bias but increases the variance (\cite{mayr12}). While \cite{bu03} considered the relative difference of mean squared errors as a stopping criterion, \cite{bu06} invoked the corrected AIC, minimizing it over a suitable set of values for $m_{iter}$. Stopping before convergence is referred to as ''early stopping'' (see \cite{bu07} and \cite{mayr12}). \cite{mayr12} complain that early stopping often requires a very high initial number of iterations and propose a strategy, based on a cross validation scheme, to circumvent this issue. 

Boosting algorithms are extremely efficiently implemented in the $\mathsf{R}$-package \texttt{mboost} (\cite{mboost}, \cite{hofner14}, \cite{hothorn10},  \cite{bu07}, \cite{hofner15}, \cite{hothorn06}).

\subsection{Stability Selection} \label{stabselsec} 

We now briefly recapitulate the Stability Selection introduced in \cite{bu10}. Let $\Lambda \subset \R_{\ge 0}$ be the set of regularization parameters $\lambda$ for Lasso-type algorithms. The estimated set of variables corresponding to a specific $\lambda$ is denoted by $\hat S(\lambda)$. When tuning the algorithm, usually by defining a grid of candidate values for $\lambda$ and fitting a model for each element of the grid, one essentially would pick one of the models which behaves best w.r.t. some quality criterion. Following \cite{bu10}, variable selection by just choosing one element of the set $ \{ \hat S(\lambda) \ | \ \lambda \in \Lambda \}$ does generally not suffice due to the overestimating behaviour of algorithms like the Lasso. Instead, one draws $B$ subsamples of a size of around $n_{sub}=\lfloor n/2 \rfloor$ and only the variables that have been selected on sufficiently many subsamples are finally selected. 

The probabilities $\hat \Pi_j(\lambda):=P(j \in \hat S(\lambda))$ are, for each $j$, approximated by computing the relative fraction of subsamples whose corresponding model contains variable $j$. Then, by fixing a \textbf{cutoff} $\pi_{thr}$, the Stability Selection defines the set \begin{equation} \label{stabsetthres} \hat S^{stab}:=\{j \ | \ \max_{\lambda \in \Lambda}(\hat \Pi_j(\lambda)) \ge \pi_{thr} \}. \end{equation} An important issue when selecting variables are type I errors, i.e., falsely selected variables. \cite[Thm. 1]{bu10} provides an upper bound for the expected number of false positives. They additionally show under which assumptions exact error control is possible with Stability Selection even in high-dimensional settings.

Since the original Stability Selection of \cite{bu10} was basically tailored to Lasso-type models, \cite{hofner15} provided a Stability Selection for Boosting models. Again, one generates $B$ subsamples and performs Boosting on each subsample. Each Boosting model is iterated until a pre-defined number $Q$ of variables is selected, respectively for each subsample. The number $Q$ and the threshold $\pi_{thr}$ (as in Eq. \ref{stabsetthres}) are related by the per-family error-rate due to \cite[Thm. 1]{bu10}, so fixing two of these quantities, the third can be reasonably set. The Stability Selection is implemented in the $\mathsf{R}$-package \texttt{stabs} (\cite{stabs}, \cite{hofner14}, \cite{mayr17}).  

An extension of the Stability Selection is introduced in \cite{shah13} who provide bounds for the type I error that are free from the exchangeability assumption and the assumption that the selection procedure is better than random guessing needed in \cite{bu10}. Another variant of the Stability Selection from \cite{bu10} has been proposed in \cite{zhou13} who criticize that the selection of the stable features is done according to $\max_{\lambda \in \Lambda}(\hat \Pi_j(\lambda))$. They suggest to take the average of the $\hat \Pi_j(\lambda)$ for the best $k$ parameters $\lambda$, calling their method therefore Top-$k$-Stability Selection. A Stability Selection including a filtering step in the sense that the grid $\Lambda$ is condensed to a sub-grid of regularization parameters corresponding to the lowest aggregated out-of-sample loss has been suggested in \cite[Sec. 8]{yu20}. \cite{yu16} proposed a method based on the so-called estimation stability which however is restricted to the choice of the regularization parameter for Lasso. \cite{pfister} suggested a stabilized regression method where regression models on different subsets of $\{1,..,p\}$ are computed and where a stability score is assigned to each model. Then, from the subset of those models with the highest stability scores, a subset of models with the highest prediction scores is taken and the corresponding models are averaged. A trimmed Stability Selection for contaminated data based on in-sample losses was introduced in \cite{TW21c}. The stability of variable selection and variable ranking has been considered in \cite{nogueira16}, \cite{nogueira17}, \cite{nogueira17b}. \cite{nogueira17b} who discuss and propose similarity metrics in order to quantify this stability.

An algorithm called Bolasso (\cite{bach08}) can be interpreted version of Stability Selection which is based on bootstrapping instead of subsampling. The final set of selected variables is the intersection of all selected sets. Therefore, one may identify the threshold $\pi_{thr}=1$ for the Bolasso.

\section{A modified, loss-guided Stability Selection} \label{ourstabselsec} 

Although the original Stability Selection already led to magnificent results in literature, there are still open questions. First, the original Stability Selection, including variants in literature, solely considers a training data set, so the stable model is completely based on in-sample losses. Second, \cite{bu} and \cite{hofner15} have made recommendations for the choice of $\pi_{thr}$ and advise not to give too much attention to it as long as it lies in a reasonable interval. On the other hand, issues with the Stability Selection as well as a considerable hyperparameter sensitivity have already been reported in literature (\cite{li13b}, \cite{wang20g}). We believe that this parameter should also be chosen data-driven by respecting the out-of-sample performance of the resulting models, analogously to cross-validation techniques to find the optimal hyperparameter from a grid of hyperparameters. Furthermore, for a user, it is much more intuitive to define the number of variables in the final model than to define a threshold since there is no intuition about how many variables a particular threshold corresponds to. This has already been suggested in \cite{zhou13}, but we also allow for a grid of such numbers so that the optimal number of stable variables is derived by the out-of-sample performance of the corresponding candidate stable models. The main feature of our aggregation and selection procedure is that it is not based on a global criterion like the PFER as in the original Stability Selection but adapted to the actual problem in the sense that the loss function directly guides the selection of the stable model.

A further problem is that the excellent implementation of the Stability Selection based on Boosting where the Boosting models are iterated until a given number of variables have been selected backfires if the hyperparameters have not been chosen appropriately since increasing this number may require to compute all Boosting models again which is rather expensive on high-dimensional data. Another problem is that, on noisy data, even the relevant variables may tend to be selected on a rather low relative part of the resamples. Therefore, it can happen that noise variables are selected more frequently than the actual relevant variables which, if one concentrates solely on keeping the number of false positives low, can lead to too sparse or even empty models.

\subsection{A grid search for the optimal stable model} \label{gridsubsec}

First, we partition the data $\D$ into a training set $\D^{train} \in \R^{n_{train} \times (p+1)}$ and a validation set $\D^{val} \in \R^{n_{val} \times (p+1)}$. The most generic procedure that we also apply in this paper is to draw $B$ subsamples $\D^{b;train}$ of size $n_{sub}$ of $\D^{train}$ and to compute predictor sets $\hat S^{(b)}$, $b=1,...,B$, by applying some model selection procedure (without any restrictions like stopping it once $Q$ variables are selected as in \cite{hofner15}) to subsample $D^{b;train}$, respectively. Then, as in literature, we compute the aggregated selection frequencies  \begin{equation} \label{hatpi} \hat \Pi_j=\frac{1}{B}\sum_{b=1}^B I(j \in \hat S^{(b)}) \end{equation} for $j=1,...,p$, and either only take all variables $j$ for which $\hat \Pi_j \ge \pi_{thr}$, or we rank the components in a descending order and take the first $q$ ones. Thus, we produce one of the final sets of selected predictors \begin{equation} \label{finalq} \hat S^{stab}(q):=\{j \ | \ \hat \Pi_j \ge \hat \Pi_{((p-q+1):p)}\}  \end{equation} or \begin{equation} \label{finalpi} \hat S^{stab}(\pi_{thr}):=\{j \ | \ \hat \Pi_j \ge \pi_{thr}\}  \end{equation} where we denote the largest element of a vector $x$ of length $p$ by $x_{(p:p)}$ and so forth.

We first decide to define the stable model either according to $q$ or to $\pi_{thr}$. In the case of adjusting $q$, we need a reasonable subset of $\N$ which clearly satisfies \begin{equation} \label{qgrid} q_{grid} \subset \{1,2,...,\#\{j \ | \ \hat \Pi_j>0\}\}. \end{equation} In the case of adjusting $\pi_{thr}$, we discretize a reasonable interval, w.l.o.g. $]0,1]$, according to some mesh size $\Delta>0$, so we get the grid \begin{equation} \label{pigrid} \pi_{grid}=\{\Delta,2\Delta,...,1-\Delta,1\} . \end{equation} Then, we search for the optimal element of the grid by first computing the candidate stable model corresponding to each grid element according to Eq. \ref{finalpi} resp. Eq. \ref{finalq}. Then, using the validation data set $\D^{val}$, we compute the final coefficient set (w.l.o.g. least squares coefficients, see Subsec. \ref{coeffsec}) for each candidate stable model and compute the loss on $\D^{val}$. This guarantees that not only the predictor sets $\hat S^{(b)}$ are adapted to the loss function but that also the stable predictor set respects the loss appropriately. See Alg. \ref{stabselalg} for an overview of our loss-guided Stability Selection.

Of course, \cite[Thm. 1]{bu10} which bounds the expected number of falsely selected variables in terms of $\bar q$, $\pi_{thr}$ and the number $p$ of variables may be applied ex post where $\bar q=\mathop{mean}_b(|\hat S^{(b)}|)$ is the average number of selected variables in each Boosting model. If $q_{opt}$ is the optimal number $q$ selected from the grid $q_{grid}$, the corresponding threshold lies in the interval $]\hat \Pi_{(p-q_{opt}):p},\hat \Pi_{(p-q_{opt}+1):p}]$ Therefore, if a grid as in Eq. \ref{qgrid} has been used, it holds (under the assumptions in \cite{bu}) that \begin{center} $ \displaystyle \E[V] \le \frac{1}{2\hat \Pi_{(p-q_{opt}):p}-1}\frac{(\bar q)^2}{p}$, \end{center} provided that $\Pi_{(p-q_{opt}):p}>0.5$. If we use a threshold grid as in Eq. \ref{pigrid} for our Stability Selection, just replace $\Pi_{(p-q_{opt}):p}$ by $\pi_{opt}$ for $\pi_{opt}$ being the optimal element in $\pi_{grid}$, provided that $\pi_{opt}>0.5$. It is important to note that we rather risk to include wrong variables in our final stable model than to get an empty or heavily underfitted model, so we do not think of not being able to compute the error control probability ex ante as a weakness of our Stability Selection. \\

\begin{footnotesize}
\begin{algorithm}[H] 
\label{stabselalg}
\textbf{Initialization:} Data $\D^{train}$, $\D^{val}$, size $n_{sub}$ of subsamples, binary variable \texttt{gridtype}, grid, hyperparameters for the underlying model selection procedure\;
\For{b=1,...,B}{
Draw a subsample $\D^{b;train} \in \R^{n_{sub} \times (p+1)}$ from $\D^{train}$\;
Apply the model selection algorithm an get a model $\hat S^{(b)}$\;
}
Compute the $\hat \Pi_j$ as in Eq. \ref{hatpi} \;
\eIf{\texttt{gridtype=='qgrid'}}{\For{$k=1,...,|q_{grid}|$}{
Get the stable model $\hat S^{stab}((q_{grid})_k)$ according to Eq. \ref{finalq}\;
Compute the coefficients on the reduced data $(X^{train}_{\cdot,\hat S^{stab}((q_{grid})_k)},Y^{train})$ and get the loss $L^{(val,k)}$ on the validation data $\D^{val}$
}}{\For{$k=1,...,|\pi_{grid}|$}{Get the stable model $\hat S^{stab}((\pi_{grid})_k)$ according to Eq. \ref{finalpi}\;
Compute the coefficients on the reduced data $(X^{train}_{\cdot, \hat S^{stab}((\pi_{grid})_k)},Y^{train})$ and get the loss $L^{(val,k)}$ on the validation data $\D^{val}$}}
Choose the model corresponding to $k_{opt}=\argmin_k(L^{(val,k)})$\;
Compute final coefficients w.r.t. the stable model on the whole data $\D$
\caption{Loss-guided Stability Selection} 
\end{algorithm} 
\end{footnotesize}

\begin{rem} \label{aggrem} The aggregation procedure may suffer from several issues, for example, contaminated instances/cells in the data set. Of course, one can apply robust model selection methods on the subsamples. Furthermore, several methods like computing the performance of the individual models on the subsamples and downweighting or trimming them when computing the $\hat \Pi_j$ have been considered in \cite{TWphd}, also including an additional outer cross-validation scheme where different partitions of the data into training and validation data are drawn. A trimming procedure based on the in-sample losses was suggested in \cite{TW21c}. \end{rem}

\begin{rem} We recommend to use $q$-grids instead of $\pi$-grids since they are more intuitive and since they guarantee that the Stability Selection does not end up in an empty model. It is important to note that the aggregated selection frequencies can clearly also be inspected, so if we force the Stability Selection to output $q$ stable variables, the analyst can investigate if they attain acceptable aggregated selection frequencies. \end{rem}

\subsection{Post Stability Selection model selection} 

Consider, for simplicity, the situation that there are $k_r$ relevant variables, indexed by $i_{k_1},...,i_{k_r}$, and $k_n$ non-relevant variables, indexed by $j_{k_1},...,j_{k_n}$, whose aggregated selection frequencies satisfy $\hat \Pi_{j_{k_1}}>\hat \Pi_{j_{k_2}}>...>\hat \Pi_{j_{k_n}}>\hat \Pi_{i_{k_1}}>\hat \Pi_{i_{k_2}}>...>\hat \Pi_{i_{k_r}}$. If the threshold is low resp. if the number $q$ of stable variables is high, one inevitably selects noise variables. These noise variables may decrease the out of sample performance, so our loss-guided Stability Selection may select only very few variables, i.e., whose aggregated selection frequencies are larger than $\hat \Pi_{j_{k_1}}$. Of course, the same argument holds for more flexible settings as above when, for example, a sequence of true variables is disturbed by single noise variable in terms of aggregated selection frequencies. Due to strictly respecting the ordering of the variables in terms of their aggregated selection frequencies, there is no chance to select the true best model if at least one relevant variable has a lower aggregated selection frequency than at least one noise variable.

We suggest to use the advantage that stable variable sets (due to the rather low threshold that we use at this stage, we term them ``meta-stable'' variable sets) are usually rather sparse. Therefore, we propose a ``Post Stability Selection model selection'' approach which combines the resampling strategy of Stability Selection with well-established classical model selection and a validation step corresponding to the loss function. We perform an exhaustive search in order to overcome the problem of inappropriate variable orderings and call the strategy `` Post Stability Selection Exhaustive Search'' (PSS-ES). See Alg. \ref{stabselexh} for an overview where $\mathcal{P}(A)$ denotes the power set of a discrete set $A$. Note that we left a definition of the ``performance'' in the algorithm open since there are several possible alternatives like standard exhaustive search w.r.t. the validation loss, additional cross-validation or a hybrid approach between the in-sample and out-of-sample performances that we apply in Sec. \ref{simsec}. 

Since the computation of the coefficients on the reduced data set is cheap, this strategy only induces an insignificant computational overhead provided that it is based on a sufficiently low number $q_0$ of variables (see Alg. \ref{stabselexh}). Alternatively, one can execute the corresponding forward or backward selection strategies. \\

\begin{footnotesize}
\begin{algorithm}[H] 
\label{stabselexh}
\textbf{Initialization:} Training data $\D^{train}$, validation data $\D^{val}$, threshold $\pi_{thr}$, maximum number $q_0$ of candidate variables, aggregated selection frequencies $\hat \Pi_j$\;
$\hat S^{meta-stab}=\{j \ | \ \hat \Pi_j \ge \pi_{thr}\}$\;
\If{$|\hat S^{meta-stab}|>q_0$}{$\hat S^{meta-stab}=\{j \ | \ \hat \Pi_j \ge \hat \Pi_{((p-q_0+1):p)}\}$}
\For{$S \in \mathcal{P}(\hat S^{meta-stab})$}{
Compute $\hat \beta^S$ on $(X^{train}_{\cdot,S},Y^{train})$\;
Compute the performance of the model}
Select the model $\hat S^{stab}$ with the best performance\;
Compute final coefficients w.r.t. the stable model on $\D$
\caption{Post Stability Selection Exhaustive Search (PSS-ES)} 
\end{algorithm} 
\end{footnotesize}

\subsection{Final coefficients} \label{coeffsec} 

We want to point out that Stability Selection aggregates models, but not coefficients. Therefore, all coefficients computed on the subsamples are not considered further. We suggest just computing the coefficients that minimize the selected loss function on the stable model, e.g., least squares coefficients for the quadratic loss function. In the presumably very rare, but not impossible case that the stable model still has more than $n$ predictors, we would apply the underlying model selection algorithm again on the reduced data set where only the stable predictors remain. Note that \cite{cher18} propose a much more sophisticated strategy for computing the final coefficients by averaging coefficients computed from solving moment equations in a cross-validation scheme which of course also could be applied here.

\subsection{Computational aspects}

As for the computational complexity of both the original and our loss-guided Stability Selection, we can state the following lemma. 

\begin{lem} \label{complem} \textbf{i)} The computational complexity of our loss-guided Stability Selection is contained in the same complexity class as the original Stability Selection provided that the complexity $\mathcal{O}(c_{base}(n,p))$ of the model selection algorithm applied on training data with $n$ observations and $p$ predictors is at least $\mathcal{O}(np)$, that the loss function can be evaluated in at most $\mathcal{O}(np)$ steps and that the final coefficients can be computed in at most $\mathcal{O}(np)$ steps. \\
\textbf{ii)} Statement i) remains valid for the Post Stability Selection model selection provided that the true number $s^0$ of relevant variables does not depend neither on $n$ nor $p$.\\
\textbf{iii)} Selecting the (candidate) stable model according to the rank-based strategy does not cause a computational overhead compared to the threshold-based strategy provided that $p$ is of order $\exp(n)$.   \end{lem}

\begin{bew} The original Stability Selection requires drawing $B$ subsamples of size $n_{sub}$, leading to a complexity of $\mathcal{O}(Bn_{sub})$, the application of the base procedure on each subsample with a total complexity of $\mathcal{O}(Bc_{base}(n_{sub},p))$, the computation of the $\hat \Pi_j$ which is done in $\mathcal{O}(Bp)$ steps and the selection of the variables according to Eq. \ref{finalpi}, requiring $\mathcal{O}(p)$ steps. Summarizing, the computational complexity is dominated by the base procedure according to the assumption, letting the total complexity be $\mathcal{O}(Bc_{base}(n,p))$. \\
\textbf{i)} The loss-guided Stability Selection with a grid of thresholds additionally requires $|\pi_{grid}|$ times to check which variables exceed the threshold, followed by the computation of the coefficients and the loss evaluation, leading to additional costs of order $\mathcal{O}(|\pi_{grid}|(p+n_{train}p+n_{val}p))=\mathcal{O}(np)$ which is captured by $\mathcal{O}(Bc_{base}(n,p))$ according to the assumptions.\\
\textbf{ii)} After having computed the meta-stable model, one additionally has to compute the coefficients and the performance measure for each considered predictor set whose number is at most $2^{q_0}$ for the exhaustive search and definitely smaller for forward or backward strategies. This again leads to an additional complexity of $\mathcal{O}(np)$. Note that this argumentation only holds if the number $q_0$ does not grow with $n$ or $p$ which is true if the assumption is valid.\\
\textbf{iii)} The only difference is that the selection of the (candidate) stable models cannot be done in $\mathcal{O}(p)$ steps due to the ordering procedure which requires $\mathcal{O}(p\ln(p))$ steps if quick sorting algorithms are applied. However, this quantity is captured by $\mathcal{O}(np)$ for $p$ being of order $\exp(n)$.  \vspace{-0.5cm} \begin{flushright} $_\Box $ \end{flushright} \end{bew}

\begin{ex} \textbf{i)} A variant of Lem. \ref{complem} clearly generally holds if the model selection procedure complexity dominates the coefficient computation and the loss evaluation. However, the $\mathcal{O}(np)$ complexity spelled out in the lemma is most natural for many coefficient estimation costs and holds for the concrete ones in this paper.\\
\textbf{ii)} It may sound counter-intuitive to require the complexity of the loss evaluation to be contained in $\mathcal{O}(np)$ since the loss requires $Y$ and $\hat Y$ as input, so $p$ does not appear. We formulated it in this way due to loss functions like ranking loss functions (see \cite{TW19b} for an overview) which require $\mathcal{O}(n\ln(n))$ steps to be evaluated. Then, assuming that $p$ is of order $\exp(n)$ maintains the assumption that the loss evaluation complexity is $\mathcal{O}(np)$. \\
\textbf{iii)}  For $L_2$-Boosting models with $m_{iter}$ Boosting iterations each, the complexity of the Stability Selection is $\mathcal{O}(Bn_{train}pm_{iter})$. \end{ex} 

\begin{rem} As for the storage complexity, the Stability Selection essentially only requires to memorize the data set and, for each $b=1,...,B$, the coefficient vector (or its logical counterpart where a TRUE appears if the coefficient is non-zero), leading to storage costs of order $\mathcal{O}(np+Bp)$. Additionally, one has to store at least the current subsample, which however could be deleted afterwards. In principle, the aggregated selection frequencies can be updated iteratively so that one only had to memorize the current coefficient vector or its logical counterpart. During the grid search, we only memorize the validation loss w.r.t. each element of the grid and finally report the coefficients of the optimal model and the aggregated selection frequencies, so denoting our grid by $*_{grid}$ for $* \in \{\pi,q\}$, the storage costs for the whole Stability Selection are given by $\mathcal{O}(np+Bp+|*_{grid}|+p)$. A Post Stability Selection model selection strategy would require the same storage capacities where $|*_{grid}|$ is replaced by $2^{q_0}$. \end{rem} 

Evidently, both Stability Selection strategies do not require any interaction between the different Bootstrap replications at any stage except for the final aggregation of the selection frequencies. Therefore, one can easily parallelize them similarly as the original Stability Selection by running the base procedure on the subsamples on different nodes.

\subsection{Other potential applications} 

We want to emphasize that our Stability Selection is not limited to Boosting models on which we focus in this paper but clearly also applicable to model selection procedures like the Lasso and its variants, as done in \cite{bu10}, where the threshold in Eq. \ref{stabsetthres} resp. a corresponding number $q$ would be selected data-driven. As in \cite{bu10}, it can also be applied to sparse covariance estimation, for example with the Graphical Lasso (\cite{banerjee08}, \cite{friedman08}). The stable model is in this case a set of nodes, i.e., a subset of $\{(i,j) \ |\ i,j=1,...,p\}$. After having computed candidate stable models, one can proceed by computing the classical covariances based on the candidate node sets and by evaluating the non-penalized log-likelihood, so the stable model with the highest value is chosen as the final model.  

A completely different setting are variable length Markov chains (VLMCs, see e.g. \cite{bu99}) where the memory length for each transition probability depends on the realizations in the previous time steps. Model selection in VLMCs consists of finding the optimal variable lengths for which \cite{rissanen} proposed the context tree algorithm where first an overfitted maximal context tree is grown which is pruned afterwards. In this sense, the final context tree can be identified with a sparse model, i.e., a sparse VLMC representation of a full MC model of the respective order. A Stability Selection would aggregate the context trees fitted on suitable subsamples of the data to get a stable VLMC representation.

\section{Applications} \label{simsec}

\subsection{Data generation}

We generate data by drawing the $n$ rows $X_i$ of the regressor matrix independently from a multivariate normal distribution $\mathcal{N}_p(M_x,I_p)$ where $I_p$ denotes the identity matrix of dimension $p \times p$ and where $M_x=(\mu_x,\mu_x,...,\mu_x) \in \R^p$ for some $\mu_x \in \R$. We fix the number $s^0$ of true relevant variables and randomly select $s^0$ variable indices from the $p$ column indices without replacement. The $s^0$ coefficients $\beta_j$ corresponding to the relevant variables are drawn independently from a $\mathcal{N}(\mu_{\beta},1)$-distribution. 

As for the signal-to-noise ratio (SNR), we can easily generate regression data with a pre-scribed SNR by computing $\Var(X\beta)$ and by adjusting the scaling of the noise term $\epsilon$ in the linear model $Y=X\beta+\epsilon$. For classification data however, we generate $X\beta$ and treat $\eta_i:=\exp(\overline{X_i \beta})/(1+\exp(\overline{X_i \beta}))$ with $\overline{X_i \beta}:=X_i\beta-mean(X\beta)$ (to avoid the issue of highly imbalanced data if (nearly) all components of $X\beta$ are positive resp. negative) as $P(Y=1)$, i.e., we draw the responses according to $Y_i \sim Bin(1,\eta_i)$. \cite[Sec. 16]{elstat} propose a noise-to-signal ratio $NSR=\Var(Y|X\beta)/\Var(X\beta)$. We are not aware of any method that allows for generating classification data with a pre-scribed SNR (or NSR) but for the sake of transparency, we compute the inverse of $NSR$ on our data sets as a replacement of the SNR.

\subsection{Training}

We compare standard Boosting ($L_2$-Boosting for regression and \texttt{LogitBoost} for classification), the loss-based Stability Selection and PSS-ES. As for Boosting, the whole data set $\D$ is used as training set. As for our Stability Selection variants, we partition $\D$ into a training set $\D^{train}$ with $n_{train}$ instances and a validation set $\D^{val}$ with $n_{val}$ instances. In Sec. \ref{artificialsec} and \ref{realsec}, we generate an independent test data set with $n_{test}$ instances that will only appear during evaluation. In all cases, in the spirit of randomized cross-validation, we draw 10 such partitions into (test,) training and validation data. The subsamples in our Stability Selection are of fixed size $n_{sub}$. Note that after having selected the final stable model, the final coefficients are computed on the whole set $\D$ using standard least squares regression resp. logit regression. The Boosting parameters are per default $m_{iter}=100$ and $\kappa=0.1$. The whole procedure is repeated $V$ times for each scenario, i.e., we generate $V$ independent data sets.

As for PSS-ES, we suggest a hybrid approach that considers both in-sample and out-of-sample losses. More precisely, due to the issue that the number of observations is often low, the number of validation samples may not be sufficiently representative so that only focusing on out-of-sample data when selecting the best candidate stable model could be misleading. Our strategy is to perform the usual exhaustive search based on the in-sample losses as implemented in the $\mathsf{R}$-packages \texttt{leaps} (\cite{leaps}) and \texttt{bestglm} (\cite{bestglm}) that compute the best model for each cardinality up to at most $q_0$ (see Alg. \ref{stabselexh}). Normally, one would select the best of these competing models by some criterion like the AIC or the BIC. In our case, we compare each of these models again by computing the out-of-sample performance of the least squares resp. logit coefficients computed only on these variables. Finally, the best (measured in the out-of-sample performance) of these best (measured in the in-sample performance) models is chosen as the final stable model. We nevertheless also implemented a forward and a backward search purely based on the out-of-sample performance and refer to them as PSS-FW resp. PSS-BW.

\subsection{Evaluation}

\subsubsection{Simulated data}

As already elucidated in \cite{wang20g}, Stability Selection is a model selection strategy. The final performance therefore depends on how the analyst uses the resulting stable model. Therefore, computing test losses would not be fair for comparing the performance of Stability Selection with that of the underlying model selection algorithm if the true model is known. 

Instead, we compute the average number of variables that have been selected for all model selection strategies. Furthermore, we compute the average number of true positives. All averages are first taken over the 10 cross-validation loops and again averaged over the $V$ repetitions. Of course, one can expect raw Boosting models to achieve a higher TPR than stable models, therefore, just reporting the number of true positives would be somewhat unfair. We compute the precision which is defined as  $pr=TP/PP$ for the number $TP$ of true positives and the number $PP$ of predicted positives, i.e., the relative part of relevant variables among the selected variables, and also report the average over the cross-validation loops and the $V$ data sets.

\subsubsection{Real data}

On real data, there is no known ground truth model. Of course, we can report the mean number of true positives, but this number alone would obviously bias the evaluation towards overfitting models. 

Therefore, we compute the losses on the test data and report the average test loss over all cross-validation samples of the given data set. The intention behind this strategy is that if the assumption of a linear model is suitable, the prediction based on the true relevant variables should be better than any prediction based on a different set of variables. Note that we compute the least squares coefficients on the selected models, including that for $L_2$-Boosting since directly using the $L_2$-Boosting output would incorporate the additional shrinking effect of that algorithm which would make the results incomparable.

\subsection{Comparison on simulated data} \label{artificialsec} 

We first compare our loss-guided Stability Selection and PSS-ES with the Stability Selection from \cite{hofner15} on three scenarios in order to highlight some issues with the latter one. Afterwards, we compare our Stability Selection variants with the corresponding Boosting algorithm on both 16 regression scenarios and 8 classification scenarios.

\subsubsection{Issues with the original Stability Selection} \label{hofnersec}

We start by showing that the Stability Selection proposed by \cite{hofner15} can fail on high-dimensional noisy data. The reason is that in such cases, there are seldom clearly dominating variables in terms of selection frequencies, or in other words, the Boosting algorithm gets irritated by the noise variables, resulting in no variable at all passing the threshold in the Stability Selection. As for Hofner's Stability Selection, we use its implementation as \texttt{stabsel} in the $\mathsf{R}$-packages \texttt{mboost} resp. \texttt{stabs}. 

One can argue that one just has to modify either $Q$, the PFER or the threshold when applying \texttt{stabsel}, but as we show in scenario A, even that does not necessarily result in good models, apart from the immense computational overhead due to increasing $Q$ necessitates to re-compute all Boosting models.

\begin{table}[H]
\begin{center} 
\begin{tabular}{|p{1.25cm}|p{0.75cm}|p{1cm}|p{1cm}|p{1cm}|p{1cm}|p{0.5cm}|p{1cm}|p{0.75cm}|p{0.75cm}|p{0.75cm}|p{0.75cm}|} \hline
Scenario&$p$ & $n_{train}$ & $n_{sub}$ & $n_{val}$ & $n_{test}$ & $s^0$ & $\SNR$ & $\mu_{\beta}$ & $\mu_X$ & $B$ & $V$ \\ \hline
A & 100 & 100 & 50 & 25 & 25 & 5 & 0.5 & 4 & -2 & 50 & 200 \\ \hline
B & 1000 & 100 & 50 & 25 & 25 & 5 & 0.25 & 0 & 0 & 50 & 100 \\ \hline
C & 1000 & 100 & 50 & 25 & 25 & 5 & 0.25 & 0 & -2 & 50 & 100 \\ \hline
\end{tabular}
\caption{Scenario specification} \label{hofnertable}
\end{center}
\end{table}

In all scenarios A-C, we apply the Stability Selection 30 times for each combination of $Q \in \{5,6,...,10\}$ and $PFER \in \{1,2,3,4,5\}$. The reason is that using one single configuration would potentially be highly misleading and discredit the original Stability Selection. Therefore, we use the test losses in order to determine the best stable model from the original Stability Selection. We do not intend to detect the best configuration here (besides, different configurations may lead to the same stable model) but to approximate the optimal model that the original Stability Selection can select.

We then only compute the number of selected variables and true positives \textbf{corresponding to the best model} in each cross-validation loop and the average over all 10 loops. We report the average of these cross-validated quantities over all $V$ replications. We also count the number of cases where no variable has been selected. The sampling type is per default the Shah-Samworth sampling type with $B$ complementary pairs. 

For the loss-guided Stability Selection, we always use a $q$-grid $q_{grid}=\{1,2,...,10\}$. We set $\pi_{thr}=0.25$ and $q_0=20$ for PSS-ES.

\begin{table}[H]
\begin{center} 
\begin{tabular}{|p{0.8cm}|p{1.25cm}|p{1cm}|p{1.35cm}|p{1.25cm}|p{1cm}|p{1.35cm}|p{1.75cm}|p{2.25cm}|} \hline
Scen.& \multicolumn{3}{c|}{Mean number of positives} & \multicolumn{3}{c|}{Mean number of TPs} & \multirow{2}{2cm}{\#\{empty models\}} & \multirow{2}{2.25cm}{\#\{all mod\-els empty\}} \\ 
& Hofner & LSS & PSS-ES & Hofner & LSS & PSS-ES & & \\ \hline
A & 3.002 & 4.71 & 5.511 & 2.783 & 2.965 & 3.102 & 0.158 & 0 \\ \hline
B & 0.672 & 3.164 & 3.39 & 0.596 & 0.909 & 0.893 & 14.012 & 11 \\ \hline
C & 0.628 & 3.222 & 3.535 & 0.547 & 0.887 & 0.902 & 15.233 & 9 \\ \hline
\end{tabular}
\caption{Results for Scenarios A-C} \label{hofnertableres}
\end{center}
\end{table}

We see in Tab. \ref{hofnertableres} that there do not seem to be severe issues with the original Stability Selection in scenario A where it achieves a significantly higher precision than our variants. Among the different hyperparameter configurations, less than 1\% of them (0.158 out of 30) led to an empty model. However, the results have to be interpreted with caution since they correspond to the best of the 30 different configurations. 

Regarding scenarios B and C where weak coefficients appear\footnote{This ``weak'' here has nothing to do with the $\SNR$ but with the magnitude of the true coefficients which has a mean of 4 in scenario A but which is centered in the other ones.} in contrast to the strong signals in scenario A, the original Stability Selection reveals that, although the best model again achieves a somewhat higher precision than our variants, it tends to severe underfitting, more precisely, around the half of the configurations (14.012 resp. 15.233 out of 30, in average) result in an empty model. Even worse, in 9 resp. 11 out of the 100 repetitions in scenario B resp. C, all configurations end up in an empty model, hence all the computational effort that has been done was in vain since on these training data sets, no variable ever entered the stable model. 

From the perspective of avoiding false positives at all costs, the original Stability Selection clearly beats our variants. From the perspective of finding a predictive model, the original Stability Selection backfires if weak coefficients appear since a considerable computational effort is lost and there is no obvious way how to proceed in this situation (would one apply the Stability Selection again with a higher false positive tolerance?). Our Stability Selection in contrast still includes false positives for the benefit that it is very easy to use without requiring expert knowledge at any stage while, when using a $q$-grid, guaranteeing to output a model to work with.

\subsubsection{Comparison of our Stability Selection with Boosting for regression}

We study the performance of our Stability Selection variants on 16 different regression scenarios, see Tab. \ref{regtable}.

\begin{table}[H]
\begin{center} 
\begin{tabular}{|p{1.5cm}|p{0.75cm}|p{1cm}|p{0.75cm}|p{0.75cm}|p{0.25cm}|p{0.75cm}|p{0.5cm}|p{0.5cm}|p{0.75cm}|p{0.75cm}|} \hline
Scenario &$p$ & $n_{train}$ & $n_{sub}$ & $n_{val}$ & $s^0$ & $\SNR$ & $\mu_{\beta}$ & $\mu_X$ & $B$ & $V$ \\ \hline
I & 1000 & 300 & 200 & 100 &  5 & 1 & 4 & -2 & 100 & 1000 \\ \hline
II & 1000 & 300 & 200 & 100 &  5 & 0.1 & 4 & -2 & 100 & 1000 \\ \hline
III & 100 & 300 & 200 & 100 &  5 & 1 & 4 & -2 & 100 & 1000 \\ \hline
IV & 100 & 300 & 200 & 100 &  5 & 0.1 & 4 & -2 & 100 & 1000 \\ \hline
V & 100 & 120 & 80 & 40 &  5 & 1 & 4 & -2 & 100 & 1000 \\ \hline
VI & 100 & 120 & 80 & 40 & 5 & 0.1 & 4 & -2 & 100 & 1000 \\ \hline
VII & 1000 & 120 & 80 & 40 & 5 & 1 & 4 & -2 & 100 & 1000 \\ \hline
VIII & 1000 & 120 & 80 & 40 &  5 & 0.1 & 4 & -2 & 100 & 1000 \\ \hline
IX-XVI & \multicolumn{10}{c|}{Identical to I-VIII but $\mu_X=\mu_{\beta}=0$} \\ \hline
\end{tabular}
\caption{Scenario specification for regression} \label{regtable}
\end{center}
\end{table}

For the loss-guided Stability Selection, we always use a $q$-grid $q_{grid}=\{1,2,...,10\}$. We set $\pi_{thr}=0.25$ and $q_0=20$ for PSS-ES resp. $q_0=50$ for PSS-FW and PSS-BW. 

\begin{figure}[H]
\begin{center}  
\includegraphics[width=6cm]{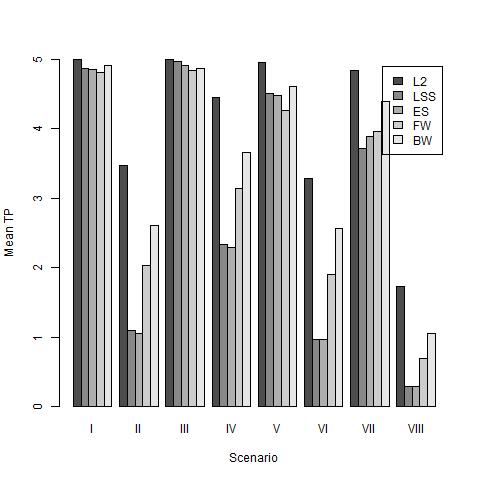}   \includegraphics[width=6cm]{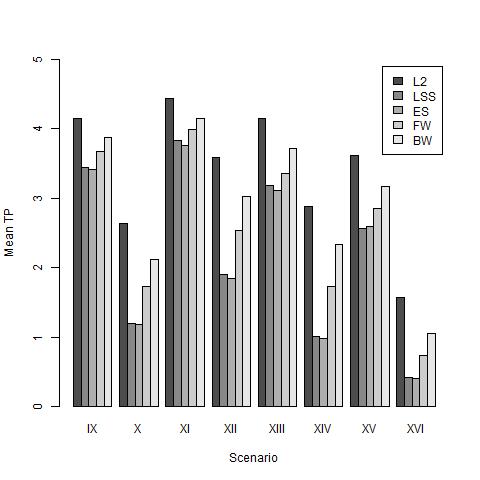}   \\
\includegraphics[width=6cm]{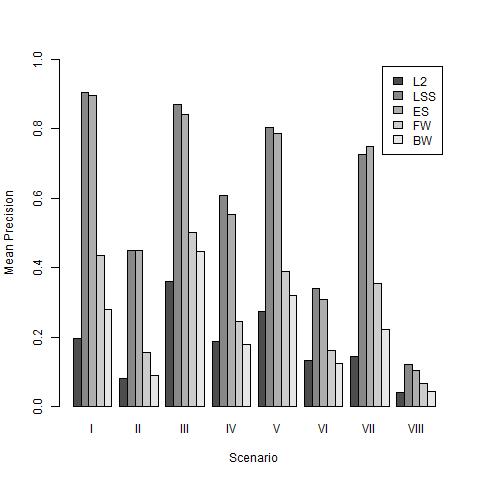} \includegraphics[width=6cm]{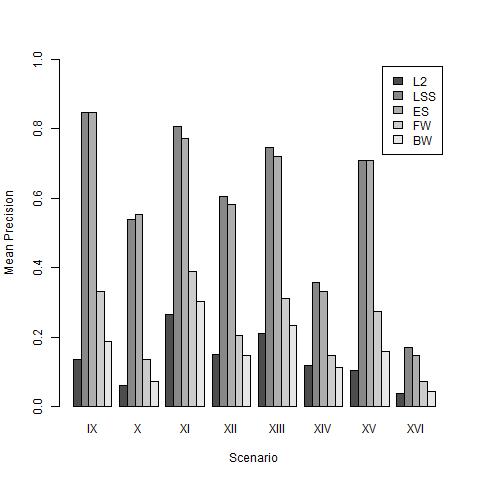}   
\caption[Stability Selection comparison on scenarios I-XVI]{Upper row: Average number of retrieved true positives for $L_2$-Boosting (L2), our loss-guided Stability Selection (LSS), PSS-ES (ES), PSS-FW (FW) and PSS-BW (BW); bottom row: Average precision}   \label{scenIXVI}
\end{center}
\end{figure}

In Fig. \ref{scenIXVI}, we can observe that our loss-guided Stability Selection and PSS-ES lead to the highest precision. More precisely, they achieve from at least double to more than five times the precision as $L_2$-Boosting in scenarios I-VIII while achieving from at least 2.8 times up to nearly 9 times the precision of $L_2$-Boosting in scenarios IX-XVI. The forward and backward strategy, although being never significantly worse than Boosting, achieve much lower precision than the other variants. We believe that this results from the rather reduced data availability due to these strategies being solely based on $\D^{val}$ while PSS-ES takes both $\D^{train}$ and $\D^{val}$ into account when finding the stable model. However, the clear dominance of the forward over the backward search can hardly be explained. One can observe a tendency of PSS-ES being slightly inferior to the loss-guided Stability Selection. 

As for the dependence of the performances on the data configuration, it is not surprising that noisy data are far more challenging than data with less noise, leading to a much lower precision on the scenarios with $\SNR=0.1$. The impact of changing $p$ does not seem to be considerable since there is no clear tendency when comparing the results of scenarios I, II, IX and X ($p$ high, $n$ high) with those of scenarios III, IV, XI and XII ($p$ low, $n$ high). Scenarios V, VI, XIII and XIV ($p$ low, $n$ low) reveal that decreasing $n$ leads to worse results while the performance on scenarios VII, VIII, XV and XVI ($p$ high, $n$ low) is significantly worse than for scenarios I, II, IX and X, so there is a clear joint impact, as expected.

It is noteworthy that the precision in the scenarios IX-XVI is comparable to that in the scenarios I-VIII although the coefficients tend to be weak while the number of true positives is lower in nearly all scenarios. Moreover, the regressors are also centered while having a mean at -2 in scenarios I-VIII. Although, mathematically spoken, the effect of the weak coefficients is just that the norms of the response and error vectors are smaller in scenarios IX-XVI than in I-VIII (in expectation), \cite{bu10} distinguish between ``active'' variables which are those with a corresponding non-zero coefficient and ``relevant'' variables whose coefficient exceeds some threshold in absolute value. However, oddly, the precision even tends to increase on the scenarios with a low $\SNR$, especially for scenarios VIII and XVI, which are the most challenging ones.

Summarizing, our loss-guided Stability Selection and PSS-ES perform well on all scenarios by achieving a drastically higher precision than standalone Boosting, without suffering from any issue like resulting in an empty model.

\subsubsection{Comparison of our Stability Selection with Boosting for classification}

We now consider 8 classification scenarios specified in Tab. \ref{classtable}. For the loss-guided Stability Selection, we always use a $q$-grid $q_{grid}=\{1,2,...,10\}$. We set $\pi_0=0.25$ and $q_0=15$ since \texttt{bestglm} does not seem to support exhaustive, forward or backward search for classification with more than 15 variables.

 \begin{table}[H]
\begin{center} 
\begin{tabular}{|p{1.5cm}|p{0.75cm}|p{1cm}|p{0.75cm}|p{0.75cm}|p{0.5cm}|p{1cm}|p{0.5cm}|p{0.5cm}|p{0.75cm}|p{0.75cm}|} \hline
Scenario&$p$ & $n_{train}$ & $n_{sub}$ & $n_{val}$ & $s^0$ & $\widehat{\SNR}$ & $\mu_{\beta}$ & $\mu_X$ & $B$ & $V$  \\ \hline
XVII & 1000 & 300 & 200 & 100 &  5 & 337.6 & 4 & -2 & 100 & 1000 \\ \hline
XVIII & 100 & 300 & 200 & 100 &  5 & 337.6 & 4 & -2 & 100 & 1000 \\ \hline
XIX & 100 & 120 & 80 & 40 & 5 & 340.4 & 4 & -2 & 100 & 1000 \\ \hline
XX & 1000 & 120 & 80 & 40 &  5 & 340.8 & 4 & -2 & 100 & 1000 \\ \hline
XXI & 1000 & 300 & 200 & 100 & 5 & 20.3 & 0 & 0 & 100 & 1000 \\ \hline
XXII & 100 & 300 & 200 & 100  & 5 & 19.7 & 0 & 0 & 100 & 1000 \\ \hline
XXIII & 100 & 120 & 80 & 40  & 5 & 19.9 & 0 & 0 & 100 & 1000 \\ \hline
XXIV & 1000 & 120 & 80 & 40 & 5 & 20.2 & 0 & 0 & 100 & 1000 \\ \hline 
\end{tabular}
\caption{Scenario specification for classification. $\widehat{\SNR}$ refers to the average of the SNR's of the individual data sets} \label{classtable}
\end{center}
\end{table}

\begin{figure}[H]
\begin{center}  
\includegraphics[width=6cm]{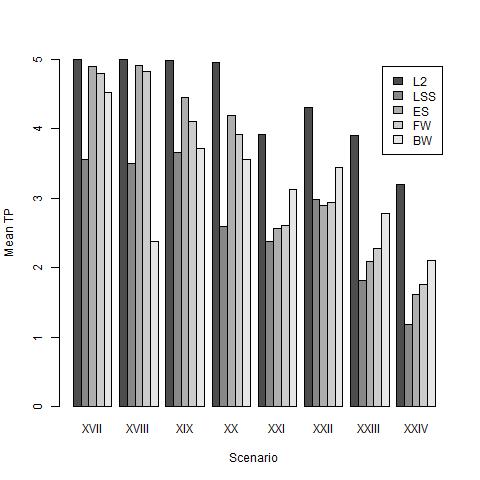}   \includegraphics[width=6cm]{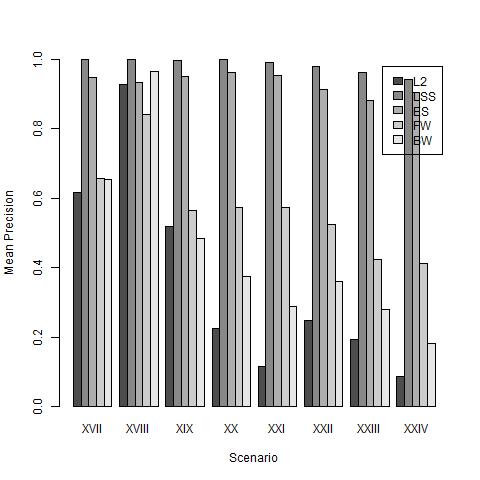}   
\caption[Stability Selection comparison on scenarios XVII-XXIV]{Upper row: Average number of retrieved true positives for $L_2$-Boosting (L2), our loss-guided Stability Selection (LSS), PSS-ES (ES), PSS-FW (FW) and PSS-BW (BW); bottom row: Average precision}   \label{scenXVIIXXIV}
\end{center}
\end{figure}  

Our loss-guided Stability Selection and PSS-ES achieve from at least the same to more than 10 times the precision as $L_2$-Boosting in scenarios XVII-XXIV. Note that the loss-guided Stability Selection achieves a precision higher than 99\% in scenarios XVII-XXI which may result from the seemingly very large $\SNR$, nevertheless, the Boosting precision is far from that value in all scenarios. 

The data configuration clearly affects the Boosting precision while the precision of the loss-guided Stability Selection and PSS-ES is nearly unaffected. The most striking difference occurs once the coefficients are weak. Besides the number of true positives also significantly decreases, there is a serious loss in the Boosting precision while the precision of the loss-guided Stability Selection and PSS-ES only decreases slightly. The PSS-FW and PSS-BW strategies again show a rather fragile behaviour. 

Summarizing, the loss-guided Stability Selection and the PSS-ES show an excellent performance on all scenarios, drastically increasing the precision in comparison to \texttt{LogitBoost}.

\subsection{Real data} \label{realsec} 

Finally, we consider the \texttt{riboflavin} data set from the package \texttt{hdi} (\cite{hdi}) which contains $n=71$ observations with $p=4088$ metric predictor variables. The response variable \texttt{y} is the log-transformed riboflavin production rate. 

We use $B=50$ (which for Hofner's Stability Selection means to sample 50 complementary pairs) and $L_2$-Boosting as base algorithm. We partition the data 100 times according to $n_{train}=50$, $n_{val}=10$ and $n_{test}=11$. For Hofner's Stability Selection, we use the default Shah-Samworth sampling scheme and apply a grid search for each of the 50 combinations of $Q \in \{11,12,...,20\}$ and $PFER \in \{1,2,3,4,5\}$ and only report the model and test loss for the best model. For our variants, we use $n_{sub}=35$ and use $\pi_{thr}=0.25$ and $q_0=50$ resp. $\pi_{thr}=0.1$ and $q_0=50$ for PSS-ES. As for the grid in the loss-guided Stability Selection, we apply different $q$-grids, namely a) $\{11,...,20\}$; b) $\{1,...,20\}$; c) $\{1,...,10\}$; d) $\{21,...,30\}$ and e) $\{1,...,30\}$. 

While these configurations serve as base scenario (call it \textbf{R1}), we also study the effect of $n_{sub}$ by setting it to $n_{sub}=25$ for our Stability Selection variants while keeping everything else as before resp. the grid as in a) (\textbf{R2}) and the effect of the partition by considering $n_{train}=40$, $n_{val}=16$, $n_{test}=15$ while keeping everything else as in R1 and while using only the grid as in a) (\textbf{R3}). 

\begin{figure}[H]
\begin{center}
\includegraphics[width=6cm]{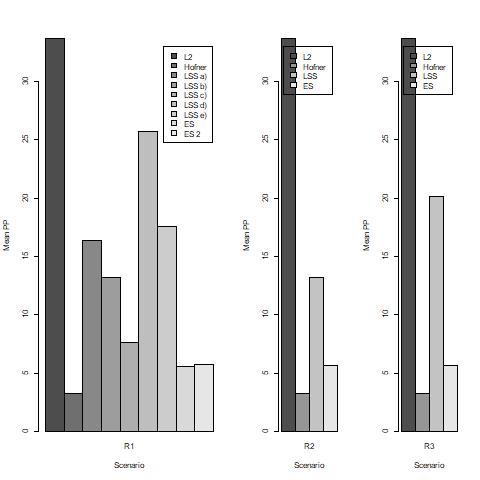}
\includegraphics[width=6cm]{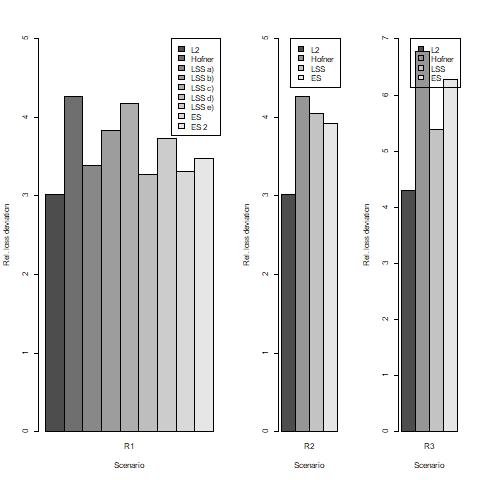}   
\end{center}
\caption[Comparison on real data]{Left: Mean number of predicted positives for $L_2$-Boosting (L2), Hofner's Stability Selection (Hofner), our loss-guided Stability Selection (LSS), PSS-ES with $\pi_{thr}=0.25$ and $q_0=50$ (ES) and PSS-ES with $\pi_{thr}=0.1$ and $q_0=50$ (ES 2); Right: Mean test losses}   \label{scenribo}
\end{figure}

In Fig. \ref{scenribo}, we depict the mean number of selected variables and the mean test losses. Note that the results for Hofner's Stability Selection and $L_2$-Boosting in scenario R1 and R2 are identical as the subsample size here only affects our Stability Selection variants. 

We can observe that, as expected, Hofner's Stability Selection leads to the sparsest models while raw $L_2$-Boosting selects the richest models. Our loss-based Stability Selection chooses quite a lot variables if the $q$-grid allows it. Of course, the mean number of selected variables in 1c) is smallest. One should note the difference between 1a) and 1b) and between 1d) and 1e). In 1a) and 1d), there is a lower bound for the number of stable variables which is not given in b) and e). Since the average number of stable variables is lower in b) resp. in e) than in a) resp. in d), there were data partitions that resulted in models with lower than 11 variables in b) and lower than 21 variables in e). Interestingly, PSS-ES selects a rather sparse stable model which is richer than the one selected by the original Stability Selection. The effect of the subsample size or the data partition seems to be rather limited concerning the number of stable variables.

Concerning the test losses, we can make the following observation. Our loss-guided Stability Selection variants achieve a better test loss than the original Stability Selection. The astounding result is that the rather sparse model selected by PSS-ES indeed achieves a much better test performance than all other stable methods, except for R1d) where the loss-guided Stability Selection outputs rather rich models which, however, only achieve a slightly better test performance. We interpret this result as follows: There are relevant variables that achieve a very high aggregated selection frequency and that enter all stable sets. There are many non-relevant variables which are however persistent and are selected on many subsamples. These variables are discarded by the original Stability Selection which cannot select the relevant variables which get even lower aggregated selection frequencies, therefore this method leads to very sparse models. Our loss-guided Stability Selection includes more variables, including these persistent noise variables, until the inclusion of these noise variables becomes infavorable concerning the validation loss. Therefore, it is capable to select much richer models than the original Stability Selection to a certain extent but is stuck to the ordering of the variables. PSS-ES in contrast reliably picks variables that, disregarding the concrete ordering in terms of aggregated selection frequencies as long as the particular frequency is at least $\pi_0$, improve the test loss. 

In this experiment, all stable models lead to a worse performance than the model selected by $L_2$-Boosting, which is not that surprising on such a high-dimensional data set since one can expect the number of relevant variables to be rather high so that stable models are doomed to underfit. In our interpretation, the false positives selected by $L_2$-Boosting have a rather low effect as the corresponding coefficients can be expected to be small in absolute value. Therefore, the rich $L_2$-Boosting model in total works better than the stable models as the benefit of the inclusion of further true variables is larger than the performance decrease due to overfitting. We want to emphasize again that an important advantage of sparse stable models is the better interpretability and the guidance of future experiments since one can focus on much less variables there.

We have also run the simulations again with $m_{iter}=300$ but this leads to such rich $L_2$-Boosting models that the number of variables is higher than the number of observations. The stable models in contrast were not affected which is not surprising since increasing the iteration number only includes some more variables while maintaining the variables selected for the first 100 iterations. The ordering of the variables would only be altered if some of these late variables would get higher aggregated selection frequencies than the early variables which is not the case and which clearly is not expected. As for PSS-ES, the number $q_0$ of meta-stable variables is quickly attained, in most cases, even for $m_{iter}=100$, so that additional variables for a larger number of iterations would just be discarded here.

\section{Conclusion and outlook} \label{outlook} 

We have shown that the standard Stability Selection where the hyperparameters are chosen manually may fail on high-dimensional noisy data due to severe underfitting, potentially caused by both inappropriate hyperparameter settings by the analyst as well as the very strict false-positive avoiding paradigm. We presented a loss-guided Stability Selection variant that chooses the hyperparameters in a data-driven way according to a grid search w.r.t. the out-of-sample loss. 

Motivated by the issue that true variables may get lower aggregated selection frequencies than some noise variables so that even our loss-guided Stability Selection cannot include these true variables without including the respective noise variables, we proposed another variant called Post Stability Selection exhaustive search (PSS-ES) which selects a meta-stable predictor set and performs an exhaustive search on this set.

The computational overhead of our Stability Selection variants is limited since the additional operations are executed on already sparse predictor sets. One can show that the complexity is captured by the same $\mathcal{O}$-class as the complexity of the original Stability Selection. Honestly speaking, the original Stability Selection may be better in the case of strong signals, i.e., high signal-to-noise ratios, since the relevant variables generally enter the models quickly.

Essentially, we do not pay too much attention to the error control but rather concentrate on finding a well-performing model rather than a too parsimonious or even an empty model which can be interpreted as a relaxed perspective in contrast to that of \cite{bu10} or \cite{hofner15}. Although we lose the error control that the original Stability Selection provided, our Stability Selection variants are user-friendly as they do not require expert knowledge, intuitive as they allow for fixing a range of number of variables in the stable model instead of some threshold, and as they, at least when using a grid over these numbers of stable variables, are guaranteed to result in a non-empty model. 

Our Stability Selection variants prove to reliably achieve somewhat higher precision, up to a factor of 10, compared to pure Boosting, in a plethora of simulated scenarios for regression and classification where the data configurations range from easy to very challenging (large number of predictors, low number of observations, weak coefficients, very large noise variance). We also applied our Stability Selection variants to a real high-dimensional data set where they prove to lead to reasonable models and where PSS-ES proves to be capable to ignore the variable ordering in terms of selection frequencies in the favor of better stable models.

\subsection{Acknowledgements} 

This paper originates from my PhD thesis at Oldenburg University under the supervision of P. Ruckdeschel but provides some different ideas and somewhat extended simulations. I thank P. Ruckdeschel for additional discussions when writing this paper, but of course, I am solely responsible for any errors.

\renewcommand\refname{References}
\bibliography{Biblio}
\bibliographystyle{abbrvnat}
\setcitestyle{authoryear,open={((},close={))}}

\end{document}